\pdfoutput=1

\documentclass[11pt]{article}

\usepackage{emnlp2021}

\usepackage{times}
\usepackage{latexsym}

\usepackage[T1]{fontenc}

\usepackage[utf8]{inputenc}
\usepackage{microtype}

\usepackage[switch]{lineno} 

\usepackage{graphicx}
\usepackage{amsmath}   
\newcommand{\nk}[1]{\textcolor{green}{Nora: #1}}

\newcommand{\eat}[1]{}
\newcommand{\red}[1]{\textcolor{red}{#1}}

\usepackage{quoting}
\newenvironment{myquote}{                   
  \parskip 0mm \begin{quoting}[vskip=0mm,leftmargin=2mm]}{
\end{quoting}}
\newenvironment{myquote2}{                   
  \parskip 1mm \begin{quoting}[vskip=0mm,leftmargin=4mm]}{
\end{quoting}}
\newenvironment{ite}{                     
     \parskip 0cm \begin{itemize} \parskip 0cm \parsep 0cm \itemsep 0cm \topsep 0cm}{
        \end{itemize}} 
\newenvironment{enu}{                   
     \parskip 0cm \begin{list}{}{\parsep 0cm \itemsep 0cm \topsep 0cm}}{
       \end{list}} 
\newenvironment{des}{                 
     \parskip 0cm \begin{list}{}{\parsep 0cm \itemsep 0cm \topsep 0cm}}{
       \end{list}} 

\title{\red{NOTE: This is an old and now obsolete draft} \\
\red{See https://arxiv.org/abs/2109.14723 for the final paper} \\
\vspace{2mm}
Enriching a Model's Notion of Belief using a Persistent Memory}

\author{Nora Kassner\textsuperscript{1,2},  Oyvind Tafjord\textsuperscript{1}, Hinrich Sch{\"u}tze\textsuperscript{2}, Peter Clark\textsuperscript{1} \\
\textsuperscript{1}Allen Institute for AI, Seattle, WA \\
\textsuperscript{2}Center for Information and Language Processing, 
LMU Munich, Germany \\
\texttt{kassner@cis.lmu.de} \\
\texttt{\{oyvindt,peterc\}@allenai.org} 
}

\begin{document}
\maketitle
\begin{abstract}
Although pretrained language models (PTLMs) have been shown to contain significant
amounts of world knowledge, they can still produce inconsistent answers to questions when probed,
even after using specialized training techniques to reduce inconsistency. As a result, it can be
hard to identify what the model actually
"believes" about the world. Our goal is to reduce this problem, so systems are 
more globally consistent and accurate in their answers. Our approach is to add a
memory component - a BeliefBank - that records a model's answers,
and two mechanisms that use it to improve consistency among beliefs. First, a reasoning
component - a weighted SAT solver - improves consistency by flipping
answers that significantly clash with others. Second, a feedback
component re-queries the model but using known beliefs as context.
We show that, in a controlled experimental setting, these two mechanisms
improve both accuracy and consistency. This is significant as it is a
first step towards endowing models with an evolving memory,
allowing them to construct a more coherent picture of the world.\footnote{Dataset is available at https://allenai.org/data/beliefbank}
\end{abstract}

\eat{
\red{Why experiments are slow:
\begin{itemize}
    \item calibration is a problem!!! When I extended the dataset the calibration seems to not work well. I changed it to a heuristic alternative which seems to work but that means I have to rerun all SAT solving experiments
    \item SAT solving takes per entity: 6 min
    \item each run of the full graph takes > 1h
    \item + 10 plants: 10 h
    \item + 20 incremental with random context: 20
    \item + 20 incremental with bm25 context: 20
    \item + 20 incremental with bm25 SAT solved context: 20
    \item + averaging over multiple splits maybe 5?: 20*4
\end{itemize}}

\red{What are the current issues:
\begin{itemize}
    \item If calibration is of F1 can still improve but it flips essential things from no-->yes and random things from yes-->no
    \item What kind of F1 should I report? Derivable facts (small test set but large improvements), gold facts (even smaller test set with good model performance and only small improvements, all IsA facts (might include a couple of wrong labels, large test set, only slight improvements as many nodes need more mutual exclusive connectivity)
    \item I have to filter plurals
    \item Do we expect the model to understand negation inside to context window?
\end{itemize}}
}

\section{Introduction}

\begin{figure}[t]
\centering
     \includegraphics[width=1\columnwidth]{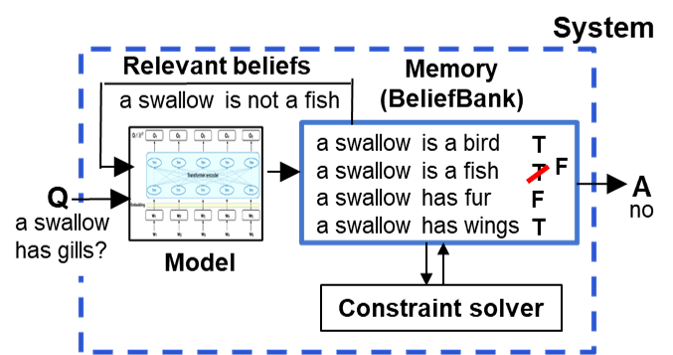}	   
\caption{ 
The proposed architecture. The model's raw answers are stored in a
persistent memory (BeliefBank), and two mechanisms attempt to improve them:
(a) A {\bf constraint solver} flips beliefs that clash significantly with others
(b) A {\bf feedback} mechanism re-queries the model about those beliefs, but
uses other, relevant beliefs as the query context. We find both consistency and accuracy
of the overall system improves. \label{architecture}}
 \vspace{-5mm}
\end{figure}

\red{NOTE: This is an old and now obsolete draft. See https://arxiv.org/abs/2109.14723
(``BeliefBank: Adding Memory to a Pre-Trained Language Model for a Systematic Notion of Belief'') for the final paper}

How might we ascribe a notion of belief to a model? Prior work has shown that, while
pretrained language models (PTLMs) contain substantial world knowledge \cite{Petroni2019LanguageMA, roberts-etal-2020-much},
their answers to probing questions can be inconsistent \cite{Elazar2021MeasuringAI, ravichander-etal-2020-systematicity, Kassner2020NegatedAM},
making it hard to pin down what a model actually ``believes'' about a proposition.
Our goal is to reduce this problem by having systems provide more globally consistent
answers to questions.

Prior work on reducing inconsistency has focused on retraining the model itself to be
more consistent in its answers, e.g., \cite{Ribeiro2019AreRR,Li2019ALF}, but with imperfect results. We present an
alternative approach in which the model is unchanged, but an evolving, persistent memory
of beliefs - called the BeliefBank - is layered on top, and two mechanisms use it 
to improve consistency among beliefs. First a reasoning component - a weighted SAT (satisfiability) solver -
flips beliefs that clash with others. Second, a feedback component re-queries the model
but using known beliefs as context, aiming for more accurate and consistent answers. 
Thus, the overall system attempts to build a more coherent representation of the
world from the model's raw answers, by ruminating on the answers seen so far.
This can also be viewed as assembling a simple ``mental model'' of the world \cite{JohnsonLaird1983MentalM} from
the noisy output of a raw PTLM.

\eat{
In addition, 
just as the model helps populate the BeliefBank, we also show how beliefs in
the BeliefBank can improve the model's answers to future questions,
by using the most relevant beliefs as additional context for question-answering
(i.e., reminding the model of important facts pertinent to the new question).
These improved answers then feed into the growing BeliefBank, allowing the
overall system (model + BeliefBank) to continuously improve performance over
time, without retraining the model itself.
}

We explore this in a controlled experimental setting where 
both candidate facts and constraints are provided. Candidate facts are simple
sentences that may be true or false, e.g., "An eagle is a bird" (T), ``An eagle is a mammal'' (F). Constraints 
are between (variabilized) facts, e.g., ``X is a bird $\rightarrow$ X has wings''. These allow us both to probe and measure improvement
in the system's consistency and accuracy. 

\eat{
\red{Delete/reword this paragraph} Our approach involves an interplay between a model's raw answers, and an
evolving memory - the BeliefBank - of beliefs based on the model's earlier answers
and a set of constraints that should hold. Model's answers contribute to the BeliefBank,
and the BeliefBank contribute new context to help with future question-answering by
the model. In between, a constraint system serves to identify and reduce inconsistency
among beliefs in the BeliefBank. Combined, this results in a system capable of
continuous improvement over time, opening new possibilities for dialog and
interactive teaching.
}

We make the following contributions:
\begin{enu}
\item[1.] We augment a PTLM with a global memory - the BeliefBank -
and show how it can be leveraged to produce more globally consistent answers.
Specifically, we compare two mechanisms for using the BeliefBank - a reasoning and a feedback mechanism - 
and demonstrate that both can improve system accuracy and consistency in a restricted setting.
\item[2.] We contribute a controlled dataset to measure a PTLM's consistency against given constraints
\item[3.] We provide an analysis of the failure modes and directions for future work
\end{enu}
These are significant as they enrich a model's notion of ``belief'',
helping them construct a more coherent picture of the world. 

\eat{
PEC: Some of these are more activities (features) rather than contributions (benefits). e.g., Proposing
an architecture isn't a contribution per se - the contribution is showing a new architecture is beneficial.

We make the following contributions: i) We propose to augment a PTLM with a global memory - the BeliefBank to track a PTLM's "beliefs" ii) We compare two mechanisms of exploiting this BeliefBank - a reasoning and a feedback mechanism iii) We contribute a controlled dataset to measure a PTLM's consistency to given constraints. iv) We show that PLMs struggle with mutual exclusive constraints. v) Finally, we show that both mechanisms improve both overall accuracy and consistency on that dataset.
}

\section{Related work \label{related-work}}


PTLMs are known to contain extensive world knowledge \cite{Petroni2019LanguageMA, roberts-etal-2020-much},
yet be inconsistent in their answers to probing questions \cite{ettinger-2020-bert,davison-etal-2019-commonsense, Kassner2020NegatedAM, ravichander-etal-2020-systematicity, Elazar2021MeasuringAI}.
While there has been some prior work on improving answer consistency, the primary approach
has been through modified model training. \citet{Ribeiro2019AreRR} improved consistency
by adding question paraphrases and question implications to the training data (data augmentation).
Others have trained models with (small) {\it sets} of examples with known constraints
between them, and included an additional loss term reflecting inconsistency among
set members during training \cite{Minervini2018AdversariallyRN,Li2019ALF,Asai2020LogicGuidedDA}.
However, the constraints are unused at test time (beyond what the model may have
internalized), and inconsistent answers are still produced.

For problems requiring a structured answer, e.g., predicting a sequence of
state changes, domain-specific constraints have been used to downgrade/block
answers that violate them \cite{Tandon2018ReasoningAA,Du2019BeCI}. This
encourages consistency within a single answer structure, but not among
different answers, our goal here.

In the area of knowledge graph construction,
\citet{Pujara2013KnowledgeGI} define ``knowledge graph identification''
as the task of building a maximally consistent knowledge graph given noisy facts
and their extraction confidences, and ontological constraints between them.
They develop a solution use probabilistic soft logic (PSL) \cite{Broecheler2010ProbabilisticSL}
as their constraint reasoner. 
Similarly, \citet{berant2010global} learn the globally
optimal set of entailments between a large database of candidate
entailment pairs (with associated confidences), by applying
a global transitivity constraint (X$\vdash$Y \& Y$\vdash$Z $\rightarrow$ X$\vdash$Z)
using Integer Logic Programming. 
In our case, we follow similar ideas but show how they can be usefully applied to the noisy predictions of a PTLM.
Specifically, we formulate the task as a weighted SAT (satisfiability) problem, and use 
the SMT solver Z3 \cite{z3} to solve it (Section~\ref{constraint-solving}).

In the area of formal knowledge-bases (KBs), efficient algorithms
have been developed for detecting, measuring, and resolving inconsistency
\cite{hansen2000probabilistic,andersen2001easy,Thimm:2009d,muino2011measuring,Thimm:2013}.
Our contribution is to leverage some of these methods for PTLMs,
adding a reasoning capability that the PTLMs alone lack.

An important part of our contribution is the use of a dynamic, persistent memory.
While there are neural architectures that include an associated memory,
e.g., \cite{Henaff2017TrackingTW,Sukhbaatar2015EndToEndMN}, these components
typically play the role of a short-term working memory to help computation.
In contrast, our BeliefBank memory layer is a persistent, long-term memory of
explicit beliefs.

Finally, our feedback mechanism uses old answers to help answer new questions.
This builds on prior work such as Self-Talk \cite{selftalk},
where a model asks itself related questions to help with new answers.
In our case, feedback is selected from a global BeliefBank, rather
than generated with templated subqueries, potentially allowing more 
control over feedback selection.

\eat{
Finally, our feedback mechanism uses earlier answers to help with new questions.
This is closely related to, and builds on, prior work where a model uses
its own answers to subquestions as 

Finally, our feedback mechanism builds on prior work where answers to

builds on prior work where a model
asks itself appropriate subquestions, and uses those answers to help
answer a main question 
}

\eat{
\subsection{Faithfulness}
\citet{subramanian-etal-2020-obtaining}
introduce the concept of module-wise faithfulness, a systematic evaluation of faithfulness in neural module networks for reasoning. We show that naive training does not produce faithful modules and propose several techniques to improve module-wise faithfulness. 
}
\eat{
\subsection{Inconsistencies in KGs}
(copied from old paper, need rewrite)
Consistency in KBs has been
studied in theoretical frameworks in the context of the
satisfiability problem and KB construction, and efficient
algorithms for detecting inconsistencies in KBs have been
proposed \cite{hansen2000probabilistic,andersen2001easy}.
Other work aims to quantify the degree to which KBs are
inconsistent and detects inconsistent statements
\cite{Thimm:2009d,muino2011measuring,Thimm:2013}.
}

\section{Task}

\subsection{Beliefs}

What does it mean to believe a proposition, say p = eagles are birds? In general, a system can
be said to (appear to) believe something if it acts as if it were true. In the specific
context of a QA system, we would expect it to produce answers consistent with p (and
its other beliefs). Pragmatically, we expect the system to (a) give a consistent answer to
different paraphrases of the question "p?" ("Are eagles birds?", "Is an eagle a type of bird?", ...),
and (b) give correct answers about implications of p ("Eagles lay eggs", "Eagles have feathers", ...).
Of course, a system may not perfectly answer such implications as the implications
may have exceptions, or the system may not be a perfect reasoner.\footnote{Similarly,
people do not always behave fully consistently with their professed beliefs.}.
Thus, to the external observer, there are {\it degrees} to which a system acts as if it
believes something.

\eat{
\red{Possibly delete this paragraph if too repetitive.} 
As has been shown elsewhere, language models (LM) can be inconsistent in their answers,
suggesting they have a rather weak notion of belief, in the sense described above.
To strengthen this, we add a dynamic memory component - the BeliefBank - on top of
the model to track and modify beliefs. We use the phrase ``the {\bf system}'' to refer
to the combined model plus belief bank. Our goal is to create a system with
a stronger notion of belief (i.e., more consistent and accurate) than the underlying LM inside it.
}

\subsection{Task Definition}

Our goal is to ascribe a stronger notion of ``belief'' to a system that includes a model $M$,
by improving the consistency and accuracy of its answers (compared with $M$). To measure
this we consider a true/false probing task, where we are also given a set of
constraints between answers:
\begin{myquote}
{\bf Given:}
\begin{ite}
\item a set of {\bf sentences} $S$ 
\item a set of {\bf constraints} $C(S)$ between (the truth values of) sentences in $S$, each annotated with a weight $w_i$ (A penalty $w_i$ is applied if $c_i \in C(S)$ is violated)
\item a {\bf model} $M$ that takes as input a True/False natural language (NL) question $Q$ and optionally an (NL) context $X$, and predicts a True/False answer $A$ with confidence score $F$
\end{ite}
{\bf Predict:}
\begin{ite}
\item the True/False labels for $S$, so as to maximally improve accuracy (with respect to gold labels) and
   consistency (minimize total penalties of constraint violations) compared with model $M$'s raw answers
\end{ite}
\end{myquote}

\section{Approach}

Our approach is to add a
memory layer, called the {\bf BeliefBank}, on top of the model to globally 
track beliefs. Two mechanisms are then used to modify BeliefBank beliefs, namely
(a) {\bf constraint reasoning} and (b) re-asking queries augmented with {\bf feedback} from the BeliefBank.

\subsection{Definitions \label{definitions}}

\begin{figure}[t]
\centering
     \includegraphics[width=1\columnwidth]{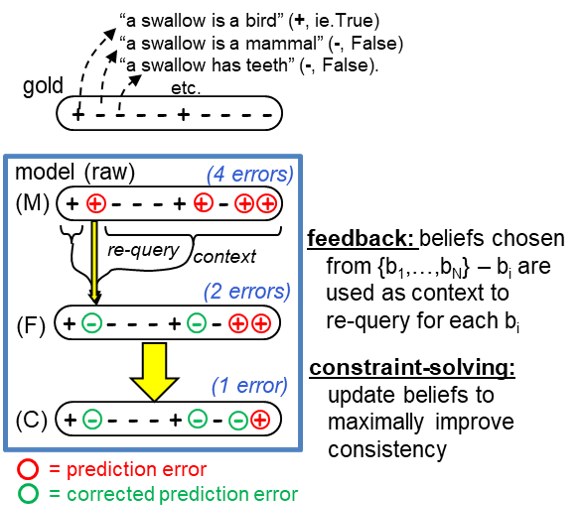}	   
\caption{Simplified illustration of iteratively improving the BeliefBank (oval). +/- denote true/false predictions for 10 facts about swallows. The model alone makes 4 prediction errors (M). Re-querying for those beliefs using other selected beliefs as context ({\bf feedback}) fixes 2 of those errors (F). Running {\bf constraint solving} on the updated BeliefBank fixes another error (C), resulting in just 1 error in the final BeliefBank. Here, the sequence is Model $\rightarrow$ Feedback $\rightarrow$ Constraints.\label{progression}}
\end{figure}

\eat{
\nk{Some comments on these definitions below: Can we find an alternative to $\textrm{tf}_i$? In our case the constraint is if X is True then Y is True/False. In general this could be also If X is False then Y is True/False. We could denote the belief as $\neg$ $s_i$/$s_i$ with an associated $w_i$ strength}

\red{Pete: I'm not quite comfortable with describing $s_i$ as an ``assertion''. An assertion is a statement
of fact or belief, but the $s_i$ themselves are just neutral sentences with no associated claims of truth or falsehood.
We could use the word ``statement'', if ``sentence'' sounds too syntactic (though as we've discussed,
the precise wording {\it is} important). Or perhaps each $s_i$ is (a sentence expressing) a statement about the world.?
Or each $s_i$ is (a sentence expressing) a possible assertion about the world?}

\red{Pete: Also, we do in fact reason with the truth values T/F of propositions, as illustrated in
Figures~\ref{architecture} and ~\ref{progression}, and in the SAT solver, so I wonder if we should call out the truth value as a 
separate entity. It seems a little strange to define $s_i \in S$ as a possible assertion, but then also say
$s_i$ is a belief. It shouldn't be both! I suppose we could say a belief is a (possibly negated) $s_i$ in the BeliefBank....}
}

\noindent Let
\vspace{-2mm}
\begin{ite}
  \item a {\bf belief} $b_i$ be a triple ($s_i$,$l_i$,$w_i$), where 
     \begin{ite}
     \item $s_i$ is a sentence $\in S$ 
      \item label $l_i \in$ \{T,F\} denotes the system's True/False belief about the truth of $s_i$
     \item weight $w_i$ is a number $\in [0,1]$ representing the system's strength of that belief
    \end{ite}
 For example:
 \begin{quote}
{\it      ("a poodle is a dog", T, 0.9) }
 \end{quote} 
 denotes the belief (strength 0.9) that "a poodle is a dog" is a true statement (T).
  \item a {\bf BeliefBank} $B(S)$ = a set of beliefs over assertions $S$ = $s_1,...,s_n$
 \item a {\bf constraint} $c_i$ = a 4-tuple of the form 
($s_i \rightarrow s_j, l_j, w_i$)
where 
\begin{ite}
\item $s_i,s_j$ are sentences $\in S$, 
\item $l_j \in$ \{T,F\} denotes the expected truth of $s_j$ if $s_i$ is true,
\item $w_i$ denotes the strength of that expectation (a penalty $w_i$ is applied if it is violated). 
\end{ite}
For convenience, we allow a shared variable X to be used in $s_i,s_j$, allowing a set of grounded
constraints to be expressed in a single statement, e.g.
\begin{myquote}
{\it (``X is a dog'' $\rightarrow$ ``X has a tail'', T, 0.8) }
\end{myquote}
expresses that if something is a dog, then it should (T) have a tail, with a penalty of 0.8 applied if
it does not. Mutual exclusivity is expressed using two rules, e.g., that fish and birds are mutually exclusive
is expressed:
\begin{myquote}
{\it (``X is a bird'' $\rightarrow$ ``X is a fish'', {\bf F}, 1.0) } \\
{\it (``X is a fish'' $\rightarrow$ ``X is a bird'', {\bf F}, 1.0) }
\end{myquote}
where ``F'' indicates the conclusion should be {\it false} if the condition is true.
\item A {\bf constraint graph} $C(S)$ = a set of constraints $c_i$ over assertions $S$ 
\end{ite}
Given a set of beliefs $B(S)$ about $S$ and a set of constraints $C(S)$, we measure
{\bf consistency} using (the complement of) \citet{Li2019ALF}'s conditional 
constraint violation ($\tau$) metric,
namely the fraction of constraints whose {\it condition} $s_i$ is believed true,
but whose {\it conclusion} (that $s_j$ has truth value $l_j$, denoted $l_j.s_j$) is not. 
In other words, over all constraints $c_i \in C(S)$, inconsistency $\tau$ is
\begin{myquote2}
$\tau = | \{ ~c_i~ |~ \neg (s_i \rightarrow l_j.s_j)~ \} | ~~~ / ~~~ | \{~ c_i ~|~ s_i ~\} |$ 
\vspace{1mm}
\end{myquote2} 
i.e., the size of the set of {\it violated} constraints ($s_i \rightarrow l_j.c_j$ is false)
divided by the size of the set of {\it applicable} constraints (i.e., those where the condition $s_i$ is true). We then define:
\begin{myquote2}
\begin{center}
consistency = 1 - $\tau$
\end{center}
\end{myquote2}

\subsection{Methods}

We evaluate two methods for improving the BeliefBank's accuracy and consistency:
\begin{des}
\item[{\bf Constraint solving:}] Given a model $M$'s answers (with confidences) in the BeliefBank, a constraint
solver seeks to reduce constraint violations by potentially flipping answers that maximally clash
with other answers. 
\item[{\bf Feedback:}] Beliefs are checked by re-querying the model, but additionally
using relevant other beliefs as context for those (re-)queries. 
\end{des}
\vspace{1mm}
Figure~\ref{architecture} shows these components, and 
Figure~\ref{progression} illustrates how they can iteratively improve the BeliefBank.
In Figure~\ref{progression}, there are 10 beliefs of interest about swallows. When probed about these, the raw model 
gets 4 of these wrong, including inconsistently believing that a swallow is both a bird and a mammal
(see (M) in Figure~\ref{progression}). Applying feedback, we re-ask the model the 10 questions but adding 
in the most relevant known beliefs as context for those (re-)queries. The BeliefBank is updated with
these new answers, fixing 2 of the errors (F). We then run the constraint solver over the BeliefBank,
resulting in one answer being flipped (question 9, from T to F), thus fixing another error (see (C)). 
The final BeliefBank has just 1 error and greater consistency.  
Note that updates can also 
introduce new errors (not shown in Figure~\ref{progression}), so improvement is not guaranteed. 

\subsubsection{Constraint Solving \label{constraint-solving}}

Given a set of beliefs and constraints, the constraint solver has two competing objectives: (a) flip 
beliefs so as to minimize constraint violations (b) don't flip beliefs, so as to preserve
the model's raw beliefs, i.e., minimize conflict between the model and BeliefBank. 
To implement this tradeoff, the model's beliefs are themselves treated as just
another constraint, e.g., the belief "a poodle is a dog" (weight 0.9) is treated as a constraint
"a poodle is a dog", with penalty 0.9 if it is violated (i.e., labeled as false by the constraint solver).
To balance the two objectives (a) and (b), the belief weights are scaled by a learned hyper-parameter $\lambda$,
trained on a calibration part our dataset, disjoint from the data then used in experiments (Section~\ref{dataset}).

To implement constraint solving, we translate the task into a {\it weighted SAT} (satisfiability) problem $P$,
for which efficient algorithms with guarantees exist. Each belief becomes a weighted assertion in $P$, e.g.,
the belief ("a poodle is a dog", T, 0.9) is expressed in SAT syntax:
\begin{myquote2}
{\it 0.9 } 
{\it "a poodle is a dog"}\footnote{In practice, strings are replaced with numeric identifiers in SAT syntax, but for clarity we leave them as strings here.}
\end{myquote2}
while the constraint ("a poodle is a dog" $\rightarrow$ "a poodle has a tail", T, 0.8) is expressed:
\begin{myquote2}
{\it 0.8 "a poodle has a tail"   ~{\bf -}"a poodle is a dog"}
\end{myquote2}
(literally: {\it "a poodle has a tail"} OR NOT ("{\bf -}") {\it "a poodle is a dog"}). 
We then apply the solver Z3 \cite{z3} to $P$, which outputs a set of truth assignments for all individual 
sentences in $P$ so as to minimize the weighted sum of violations. If the truth assignment of any sentence
has changed, the BeliefBank is correspondingly updated. 


\subsubsection{Feedback}

Feedback involves re-asking the model a question, but with the benefit
of knowing answers to related questions. To use these answers in the re-query,
selected beliefs are added to the query context before re-asking the model.
(Note that the selected beliefs are not guaranteed to be correct, of course).
Our conjecture is that if the model is explicitly reminded of relevant beliefs when answering a new question,
it will answer the question more accurately and consistently.
For example, in Figure~\ref{architecture}, when asked "Do swallows have gills?", our model $M$ incorrectly answers "yes".
But if reminded that swallows are not fish, by asking: "CONTEXT Swallows are not fish. QUERY Do swallows have gills?" 
the model now correctly answers "no". 

We evaluate two policies for choosing which beliefs to feed back to $M$ when re-asking question $Q$:
\begin{enu}
\item[1.] randomly selected from the BeliefBank
\item[2.] most relevant, using the constraint graph to identify relevance. As the constraint graph captures potential clashes that the answer to  $Q$ could cause, we use the graph to select those beliefs that would be most affected by that answer. 
For example, if the current query is: "Is a poodle an animal?", the constraint graph identifies potential clashes the would occur if the model
answered "yes", and also clashes if it answered "no". In this example, if the model answered "no", the resulting belief ("a poodle is {\it not} an animal") would strongly clash with other beliefs "A poodle is a dog.", "A poodle is a mammal.", and "A poodle is a domesticated canine.", so all three are strong candidates for the context. We select the three overall strongest clashing beliefs found in this way, considering both a "yes" and a "no" answer to $Q$.
\end{enu}
In both cases, three (if possible) beliefs are selected, this number empirically found to be most effective. (For experiments with small amounts of data, sometimes less than three relevant beliefs are found.)

\section{Dataset \label{dataset}}


We create a dataset to test our approach in a controlled way, allowing us to perform systematic experiments to evaluate behavior.
The dataset contains two parts, {\it constraints} and {\it facts}, defined over simple sentences
such as ``a swallow is a bird.''.

\subsection{Constraints}

The dataset contains two kinds of constraints:
\begin{des}
\item[{\bf positive implications:}] (conclusion truth value $\textrm{l}_i$ = True), e.g.,
\begin{myquote}
{\it ``X is a dog $\rightarrow$ [TRUE] X has a tail.''}
\end{myquote}
\item[{\bf mutual exclusivities:}] expressed as a pair of constraints with $\textrm{l}_i$ = False, e.g.,
\begin{myquote}
{\it ``X is a dog $\rightarrow$ [$\textrm{l}$=FALSE] X is a bird.''} \\
{\it ``X is a bird $\rightarrow$ [$\textrm{l}$=FALSE] X is a dog.''}
\end{myquote}
expresses that an entity cannot be both a dog and a bird at the same time.

\end{des}
Positive implications were manually gathered from ConceptNet \cite{Speer2017ConceptNet5A}.
First, we identified 176 general concepts of interest, e.g., ``mammal'', choosing concepts
with high occurrence ($>$ 100 times) in ConceptNet, avoiding significantly ambiguous terms (e.g., ``bat''), and filtering out plurals and 
obscure concepts.
For these entities, we then collected all ConceptNet facts involving 6 relations: IsA, HasA, MadeOf, PartOf,
HasProperty, and CapableOf, and
re-expressed them as constraints. For example, the ConceptNet triple (dog, HasA, tail) gives rise to the constraint
"X is a dog" $\rightarrow$ "X has a tail." 
(Triples are converted into English sentences using simple templates).
We then manually filter theses constraints for factual correctness.
We also add the constraint in the backwards direction, "X has a tail" $\rightarrow$ "X is a dog",
expecting these to have lower weight, i.e., be weaker.
(These backwards rules discourage the trivial solution that everything is false.)
Finally, weights are assigned to
all the constraints using a combination of crowdsourcing and calibration,
described shortly (Section~\ref{section:penalties}).

\eat{
resulting in 
to exclude wrong implications like \red{XXX}. Finally, we use crowdsourceing to validate these constraints, see \ref{section:penalties}.
}
\eat{Mutual exclusivities were gathered from a filtered subset of the WordNet taxonomy,
based on the approximation that (generally) siblings in the WordNet noun hierarchy
are mutually exclusive \red{elaborate}.
}

Mutual exclusivities were gathered from the ``isa'' taxonomies in ConceptNet and WordNet \cite{wordnet},
using the approximation that (generally) siblings in the noun hierarchy are mutually exclusive.
Thus, for any pair of concepts in our concepts of interest that reside in different taxonomic subtrees,
we add a mutual exclusivity constraint (using two constraint rules). 

We collected 12,147 constraints in this fashion (1798 implications, 8599 mutual
exclusivities). The set of constraints can be 
viewed as a {\it constraint graph} of implications from one sentence to another.

\subsection{Constraint Weights}
\label{section:penalties}
We used a combination of crowdsourcing and calibration to assign a reasonable
constraint weight $w_i$ to each constraint. Workers were shown each
constraint and asked to judge if the implication held always/usually/sometimes/rarely/never with raw scores 
4/3/2/1/0. 
Three workers independently scored each constraint and the scores averaged.

When used with a particular model $M$, the raw scores need to be recalibrated
to appropriately match the confidences output by that model. This is described shortly in Section~\ref{calibration}.

\subsection{Facts}

We also collect a set of truth-labeled facts about different entities, relevant to the constraints.
To do this, we select a new entity, e.g., "poodle", that is a member of one of our general concepts,
e.g., "dog", then instantiate the constraint graph with that
entity (i.e., set X = "poodle"). We then identify the leaf (source) nodes of that graph, just considering
forward implication rules, i.e., find facts not implied by other facts in the graph, and manually
annotate their True/False labels. We then use the implications to infer
other True/False labels for other sentences, i.e., propagate the annotated labels through the graph.
This provides ``silver'' labels for sentences reachable in this way (a subset of all the sentences 
in the graph), silver because the implications are soft, hence not guaranteed to hold for all entities. 

We repeat this for 27 entities (17 animals, 10 plants), resulting
a final dataset containing 4998 ``silver'' facts (sentences + True/False labels).
\eat{Note that each slice is independent and can be processed separately.
As described shortly, we perform our experiments with each slice in turn
and average results.}

\eat{
\subsection{Query and sentence templates}
\nk{Not sure where to put this section}
In order to query the model for its belief we manually define template queries for each relation, e.g., for the IsA relation the template query "Is X a Y?" is used.
These template queries are then instantiated for each sentence of the constraint set by filling in the relevant objects Y, e.g., "Is X a dog?".

We then query the model's beliefs involving an entities of the test set by filling in the subject slot X, e.g., "Is a poodle a dog?".

We also define an additional set of sentence templates, e.g., "X is Y." which are used to feed back beliefs into the context window of the model.
}

\eat{\subsection{Templates}

We collect all the unique variabilized sentences in the constraints together,
resulting in 200 \red{??} sentence templates, e.g., ``?X is a dog.''.
}

\begin{table*}
\centering
{\small
\begin{tabular}{|l|ll|ll|ll|} \hline
Results after seeing \% of data $\rightarrow$ & \multicolumn{2}{|c|}{\bf 10\%} &
\multicolumn{2}{|c|}{\bf 50\%} &
\multicolumn{2}{|c|}{\bf 100\%} \\
Accuracy (F1), Consistency (1-$\tau$) $\rightarrow$ & F1 & Con & F1 & Con & F1 & Con\\ \hline           
model (raw) & 75.07 & 87.06 & 74.29 & 77.11 & 73.25 & 74.68 \\
\eat{model + constraint-solving & 80.2 & 100.0& 88.72 & 99.81 & 88.42 & 99.60 \\}
model + constraint-solving & 81.23 & 100.0 & 89.80 & 99.93 & 91.90 & 99.83 \\
model + feedback (random)& 85.33 & 93.46 & 84.59 & 87.59 & 84.00 &88.02  \\
model + feedback (using constraints)& 82.04 & 94.69 &84.74&92.24&88.28&91.00 \\
model + feedback (using constraints) + constraint-solving & 83.69 & 100.0& 91.66 & 99.53 & 93.40 & 99.42\\ \hline
\end{tabular}
}
\caption{Results after seeing different proportions of data.
{\bf model (raw)} are scores for the model stand-alone.
In {\bf model + constraint-solving}, the constraint solver is run over all
the raw answers so far. In {\bf model + feedback}, questions are re-asked
using context selected from the other raw answers so far. In {\bf model + feedback + constraint-solving},
the constraint solver is additionally run on the answers with feedback. 
\label{results}}
\end{table*}

\section{Experiment Environment}

\subsection{Model}

The fixed model $M$ that we use for our experiments is the multi-angle QA system MAQAw \cite{arc-da}.
MAQAw is a derivative of UnifiedQA, a state-of-the-art T5 model fine-tuned on $\approx$400k question-answer pairs
\cite{unifiedqa}. MAQAw was then 
further fine-tuned on several thousand science questions and trained on different permutations of inputs and outputs
(e.g., query Q + context X $\rightarrow$ answer A; XA $\rightarrow$ Q; etc.). 
To query the model's beliefs we pose the query and let the model chose between the two answer options "yes" and "no".
MAQAw also outputs an answer confidence, used as the belief weight. Note that we do not retrain MAQAw for
this work; rather, it is used as a black-box QA module in the broader system (Figure~\ref{architecture}).

\subsection{Calibration \label{calibration}}

To appropriately mix MAQAw's confidences and the constraint scores, a calibration step is needed. To do this,
we use the 1239 facts about seven entities in our dataset as a calibration set, and then perform
experiments using (only) the remaining facts. 

We calibrate the crowdsourced raw constraint scores using sigmoid-scaling (two parameters).
We also calibrate the relative weight $\lambda$ of the model's beliefs and the constraints,
the relative weight of the backward implications (compared with the forward), and 
the relative weight of mutual exclusivity rules.
To do this we perform a grid search over these parameters, finding the values that result in
the highest F1 (accuracy) after running the constraint solver over the raw model's beliefs about
these facts.

\section{Experiments}

We evaluate accuracy (F1)\footnote{
We measure accuracy with F1 (on the True class) rather than \% correct because
the True/False distribution in our dataset is unbalanced, with significantly fewer True than False
answers. F1 avoids scores being dominated by negative answers.}
and consistency (1-$\tau$, Section~\ref{definitions}) 
for 3 different-sized, randomly selected subsets of our data: 10\%, 50\%, 100\%. We average results across entities.
The results are shown in Table~\ref{results},
showing the model's a priori accuracy and consistency (line 1),
and the effects of constraint-solving, feedback, or both (lines 2-5).

\subsection{Results}

\noindent
Several conclusions can be drawn: 
\begin{enumerate}
\item {\bf The basic model is both imprecise and inconsistent,} with F1 $\approx$ 73\% and consistency $\approx$ 75\%.
This reflects similar observations of model inconsistency made for other PTLMs (Section~\ref{related-work}), and illustrates the
challenge we wish to address. As we show shortly, the main source of error is in precision, rather than recall,
i.e., the large majority of errors are false positives.

\item {\bf Constraint-solving removes almost all inconsistency} with respect to our constraints, even with
only a subset of the data. (Of course, with less data, there are fewer applicable constraints that can be
violated.) It also {\bf significantly improves accuracy} ($\approx$+10\% F1), indicating
that the flipped truth values not only reduce clashes but better align with the true (gold) labels. Note
that improvement in accuracy is not guaranteed: constraint-solving can (and sometimes does) flip
truth values the wrong way, in order to reduce constraint violations. The improvement suggests 
that the model is getting enough answers correct, a priori, to steer the constraint-solving in
the right direction. Comparing accuracy after seeing 10\%, 50\% and 100\% of the data, we note incremental gains the more beliefs and constraints the constraint-solver is able to leverage.   

\item {\bf Feedback (both types) also improves both accuracy and consistency}, although not
to the same extent as constraint-solving. It is perhaps surprising that explicitly reminding the model
of facts it already believes, when answering a new question, can improve the results.
The context may be behaving loosely like an attention mechanism, encouraging the
model to focus on the facts provided in the context, even if they are already latently known,
thus influencing question-answering behavior. Similar observations have been observed by \citet{selftalk}. 

\item {\bf Constraint-guided feedback is more effective than random feedback} when the BeliefBank is
large (100\% column), with F1 rising to 88\% and consistency to 91\%. 
Conversely, 
randomly selected feedback perform similarly across different data sizes. We analyze this further in Section~\ref{analysis}.

\item {\bf Combining constraint-guided feedback with constraint solving further improves accuracy} by $\approx$5 percent points compared with only using feedback,
and by $\approx$2 percent points compared with the constraint-solver alone.
Again the addition of the constraint-solver results in
almost perfect consistency of 99\%. This setup reflects the incremental setup depicted in Figure \ref{progression}, where 
the constraint-solver can leverage the updated BeliefBank.

\end{enumerate}

\subsection{Analysis \label{analysis}}

Both constraint-solving and feedback can cause the overall system to flip beliefs, with
the goal that the overall system's beliefs (in the BeliefBank) are more accurate and consistent
than those of the underlying model. However, both mechanisms have the potential to flip
beliefs in the wrong direction also. We provide some analysis of both good and bad flips
here.

\paragraph{The System Behaving as Desired:} 
As an illustration of desired behavior, the raw model incorrectly believes that a pine is both a plant (correct) and a
vertebrate (incorrect), when queried. However, this violates a mutual exclusivity rule, so the constraint-solver
considers flipping one of these. Flipping ``pine is a plant'' from T to F would result in 
numerous other violations, e.g., ``pine is a tree'' (which the model also believes)
would become violated. As a result, it prefers to (correctly) disbelieve ``pine is a vertebrate'', improving
both the accuracy and consistency of the BeliefBank.

\paragraph{Types of Error:} From an analysis of the data, we see that the majority of the raw model errors
are false positives - the MAQAw model generally answers (almost) all the positive facts correctly (recall is $\approx$97\%),
but mistakenly thinks some negative facts are also true (precision is $\approx$60\%).
These false positives are typically rather unusual facts, e.g., ``A poodle is a bathroom.'' (MAQAw's answer: True).  
It is unsurprising that the model knows most of the positive facts, as they are simple statements about 
common entities (``eagles can fly''), likely seen in pre-training. However, the fact that the model 
makes (what a person would view as) catastrophic errors when asked more unusual
questions, e.g., believing that ``a poodle is made of fermented milk and bacteria'', 
reveals that the PTLM's grasp of the world is still incomplete and problematic. The constraint mechanism
proposed here essentially asks the model to think about its answers and their consequences,
so that it can spot problems that the PTLM alone does not see, and repair them accordingly.

\paragraph{Sensitivity to Constraints and Weights:} 
The constraint reasoner also makes mistakes, sometimes flipping things the wrong way
so as to improve consistency, at the expense of accuracy. For example, the raw model
correctly believes that a ``a rat is not a cat''. 
However, the constraint solver then (incorrectly) flips this to "a rat is a cat". It does this as
there are multiple constraint rules weakly suggesting rats are cats given the model's other beliefs
(''rats catch mice'', ``rats have tails'', ``rats have fur'',...), which together add up. However,
the model {\it also} (correctly) believes ``a rat is not a feline'', and there is a constraint
that ``a cat is a feline'', so in principle this should prevent the belief ``a rat is a cat''.
In practice, though, the constraint ``a cat is a feline" does not have infinite weight,
so here the constraint mechanism allows it to be violated, allowing the wrong
conclusion (``a rat is a cat'') to be drawn.

Of course, one could increase the weight on the ``a cat is a feline'' constraint
to solve this particular problem, or add new constraint rules like ``cats are larger than squirrels''. In general, though,
the effectiveness of constraint reasoning is sensitive to both the number
and weights on the constraints. Although we have used automated techniques
to re-calibrate the various weights (Section~\ref{calibration}), the system's
behavior remains sensitive to them. Finding more effective ways to discover and
appropriately tune constraints remains an open problem.

\paragraph{Ambiguity:} Although we have avoided obvious cases of ambiguity, e.g., reasoning
about ``bats'', there are smaller cases of ambiguity that may explain some of the raw
model's errors. For example, although the model (incorrectly) believes "a swallow is a fish",
there {\it is} a fish called "a black swallower" that it may have seen in pre-training,
confusing it. Similarly, some unusual statements such as ``A penguin is a protein''
(gold label is False, model believes is True) are partially ambiguous, possibly
contributing to the discrepancy.

\paragraph{Does the Model Know the Constraints?} 
Finally, we are checking consistency using {\it externally} defined constraints,
which the raw model may not itself be aware of (i.e., may not have acquired in pre-training). 
For example, although the model may {\it appear} to be internally inconsistent if it thinks a swallow is both a bird 
and a fish, this is only true if it also knows these beliefs are mutually exclusive.
To check for {\it internal} inconsistency of a model's behaviour, we would also need to
check if the model knew the violated constraints themselves, e.g., using additional probes.
Again this is an area for future exploration.

\paragraph{The influence of Feedback:} We currently can only speculate why feedback (i.e., providing relevant
facts to the model as context for question-answering) improves results. One explanation is that
feedback helps the model focus attention. For example, reminding the model that "a swallow is not
a fish" should help it realize that "a swallow has gills" is False (Figure~\ref{architecture}).
However, we also observe significant gains when feeding back random facts about the entity
being queried. Possibly these facts still encourage the model to focus on the entity. Or,
given the majority of beliefs are negative, the simple presence of negation in the context
may discourage false positive answers, the primary cause of raw model errors.

\section{Future Work}


These findings were made using a restricted experimental setup where facts are short sentences, constraints are provided, and both facts and constraints use the same language so it is clear when constraints apply. To broaden the approach, several developments are required:
\begin{ite}
\item
A system would need to {\bf automatically gather constraints and queries}.
Constraints might be mined from text, or even extracted from the model itself by direct querying. Alternatively, domain-specific constraints (e.g., ``X is a bird $\rightarrow$ X has wings'') could be replaced with more generally applicable constraint patterns
(e.g., ``X is a Y \& Y has Z $\rightarrow$ X has Z''), turning the domain-specific parts of constraints into facts that
themselves could reside in the BeliefBank.
\item Our queries have been simple facts. To fully utilize the BeliefBank for more complex
queries (e.g., ``What color is the gas that plants produce?''), a mechanism would be needed to {\bf decompose compound queries
into their primitives} (``What gas do plants produce?'', ``What color is that gas?''), and ensure they were
consistent with BeliefBank beliefs (e.g., ``Oxygen is colorless''); in other words, ensure complex answers were {\it faithful} 
to the BeliefBank \cite{Subramanian2020ObtainingFI}.
\item In general, detecting when a constraint applies to a fact is non-trivial, and requires appropriate machinery for
{\bf identifying sentence/constraint alignment}, e.g., an entailment model \cite{Seo2017BidirectionalAF}.
\item Although constraint/model weights are automatically calibrated (Section~\ref{calibration}), the
system is still sensitive to those weights and can make mistakes (Section~\ref{analysis}). {\bf Improved mechanisms for automatic calibration}
would significantly help.
\item Our system is passive, in that it answers questions and uses answers in the order they are provided. An alternative design
would be an {\bf active system} that, given a new question, actively identifies which auxiliary questions would be most
useful to also ask, to help answer the main question (either with constraint solving or feedback).
\end{ite}
Finally, scaling to a massive BeliefBank would require addressing several engineering requirements,
including bounding the constraint checking, and not re-asking every prior question as new answers were discovered.
For example, we might only check constraints
and their implications up to depth $D$ from a new fact, rather than exhaustively. Similarly, the system may 
maintain a BeliefBank where earlier answers were generated using only feedback available at that point, rather than re-asking
all earlier questions as new knowledge becomes available.

\eat{
\red{Mention active (targetted) question-answering by the system, rather than passive processing of an external stream of probes}
Despite these, the concept of augmenting a model with a memory layer plus reasoning is generally applicable,
allowing both neural and structured reasoning to interact. This may serve as a useful, extended architecture
for future systems.
}
\eat{
These findings are based on a very controlled experimental setup: constraint gathering is resource-bound, e.g., ConceptNET and not straight forwardly applicable to new domains. The system is based on carefully calibrated penalty scores.

In order to scale to real QA several modifications have to follow: Ideally future work will develop a system that handles queries beyond simple knowledge base triplets and is able to automatically gathers relevant constraints and queries.
}
\eat{
\begin{ite}
\item Beyond triples
\item Sources of constraints
\item Source of queries (e.g., which questions should we ask)
\item Resource-bounded constraint solving (e.g., explore just to depth D)
\end{ite}
}

\section{Conclusions}

PTLMs can be inconsistent in their answers to probing questions, and
can still give (what to a person appears as) naively wrong answers to
unusual questions. This work is a first step to alleviating these
problems.  
By augmenting a PTLM with a persistent, global memory - the Belief Bank -
and using it for both constraint-solving and feedback, we have shown
that both consistency and accuracy can be significantly improved.
The additional memory layer can loosely be seen as a simple ``mental model'',
constructed from the PTLM's raw answers.

Our experiments were conducted in a restricted, controlled setting,
and further development is needed to scale to larger and more
complex tasks. Nevertheless, the work here is significant as it
is a first step towards endowing models with 
a globally consistent notion of belief,
allowing them to construct a more coherent picture of the world.
The dataset is available at https://allenai.org/data/beliefbank.


\bibliography{custom, anthology}
\bibliographystyle{acl_natbib}

\end{document}